\pdfoutput=1
\documentclass[runningheads]{llncs}

\usepackage{times}
\usepackage{latexsym}
\usepackage{fvextra}
\usepackage[numbers]{natbib}
\usepackage{float}
\usepackage{enumitem}

\usepackage[T1]{fontenc}
\usepackage[utf8]{inputenc}

\usepackage{microtype}

\usepackage{inconsolata}

\usepackage{graphicx}

\title{Evaluating Large Language Models for Diacritic Restoration in Romanian Texts: A Comparative Study}
\titlerunning{LLMs for Romanian Diacritic Restoration}

\author{Mihai Dan NADĂȘ\inst{1} \and Laura DIOȘAN\inst{1}}
\institute{Babeș-Bolyai University, Cluj-Napoca, Romania\\
\email{mihai.nadas@ubbcluj.ro, laura.diosan@ubbcluj.ro}}

\begin{document}
\maketitle
\begin{abstract}
Automatic diacritic restoration is crucial for text processing in languages with rich diacritical marks, such as Romanian. This study evaluates the performance of various large language models (LLMs) in restoring diacritics in Romanian texts.

Utilizing a comprehensive corpus, we tested models including OpenAI's GPT-3.5, GPT-4, and GPT-4o, Google's Gemini 1.0 Pro, Meta's Llama 2 and 3, MistralAI's Mixtral 8x7B Instruct, Deepinfra's airoboros 70B, and OpenLLM-Ro's RoLlama 2 7B, across different prompt templates ranging from zero-shot to complex multi-shot instructions. Our findings indicate that models such as OpenAI's GPT-4o achieve high diacritic restoration accuracy, significantly surpassing a baseline echo model. However, other models, specifically those from Meta's Llama family, showed varied performance, highlighting the impact of model architecture and training data on task-specific outcomes.

This research underscores the need for specialized finetuning and model enhancements to improve NLP tasks involving diacritic-rich languages, providing valuable insights for future developments in computational linguistics.
\end{abstract}

\section{Introduction}
Romanian, known for its complex diacritical landscape, often sees these important marks disappear in digital texts. This usually happens because of limitations in keyboard layouts and errors during data entry. When diacritical marks are lost, it introduces ambiguities and significantly reduces the quality of text processing. This issue is particularly relevant today, given the advent of novel Generative AI models. For these models to produce high-quality results, they rely on training with high volumes of high-quality data. Hence, the accuracy and integrity of the input data, including the proper use of diacritics, becomes crucial.

This study aims to bridge this gap by empirically evaluating various LLMs for diacritics restoration in Romanian text. We adopt a novel approach, experimenting with different prompt templates in LLMs, informed by the effective approaches to attention-based neural machine translation.

The intrigue around LLMs comes from their remarkable versatility in handling a diverse range of tasks, including those that involve considerable variability in input data. Although RNNs, CNNs, and NMT-based methods have shown high accuracy on specific NLP tasks, including diacritics restoration, the overhead associated with fine-tuning input data and subsequent training phases may render these approaches less feasible than utilizing LLMs directly with raw text and specific prompts.

The findings are significant, revealing disparities in LLM performance, some even underperforming against the neutral baseline. This not only highlights the current capabilities and limitations of LLMs in handling diacritical marks, but also offers insights for future advancements in the field.

\subsection{Research Questions}
Our investigation focuses on several key questions:

\begin{itemize}
    \item RQ1. To what extent can LLMs, when applied with various prompt templates, effectively restore diacritics in Romanian texts?
    \item RQ2. How do the performances of these models compare against a neutral baseline, particularly in terms of Restoration Accuracy and Restoration Error Rates?
    \item RQ3. How do the results of LLMs vary in the correction of diacritics, and what factors, such as model architecture or language complexity, contribute to these differences?
\end{itemize}

Through the lens of these questions, our study endeavors to shed light on the nuanced dynamics of diacritics restoration, thereby contributing to the ongoing discourse on enhancing NLP methodologies for linguistically diverse contexts.

\section{Background and related work}
Diacritics, essential in conveying meanings and nuances, play a critical role in various languages, as highlighted in \cite{naplava2018diacritics}, 
where they are integral to the correct representation of phonetic and grammatical aspects.

\subsection{Diacritics in Natural Language Processing}
The restoration of diacritics in text processing, a multifaceted problem as described by \cite{laki2020automatic}, involves understanding context, syntax, and semantics at multiple levels. For languages such as Romanian, which employs a range of diacritical marks, the absence of these marks can lead to substantial ambiguities. This restoration process is both a technical and linguistic challenge, requiring deep understanding of the language's nuances.

In the exploration of automatic diacritics restoration for Romanian, the study by Nuțu et al. \citep{nuctu2019deep} presents a comparative analysis of six neural network architectures that integrate both recurrent and convolutional layers, without relying on additional linguistic or semantic input. Their approach, focusing solely on character sequences, demonstrates the effectiveness of deep learning strategies in enhancing the accuracy of diacritics restoration. The best performing model in their experiments, a CNN-based architecture (\textit{seq2seq\_CNN}), achieved notable accuracy levels of 97\% at the word level and 89\% at the diacritic level. This work underscores the potential of employing advanced neural network models to address the complexities associated with the restoration of diacritics in Romanian text, providing a significant contribution to the field and setting a foundation for future research endeavors that might explore the integration of Large Language Models (LLMs) for this purpose.

\subsubsection{Existing datasets and characteristics}
For the purpose of diacritic restoration in Romanian, one notable dataset is a subset of the CoRoLa text corpus, as utilized in Nutu et al.'s \citep{nuctu2019deep} study. This subset comprises 51,043 sentences, encompassing over 1 million tokens and 63,194 unique words, predominantly from belletristic texts. This dataset is of particular relevance due to its comprehensive coverage of contemporary Romanian language usage, providing a reliable source of correctly diacritized text, which is crucial for training effective models.

The corpus is not specifically designed for Automatic Diacritics Restoration (ADR) tasks but is manually annotated at the word level with various linguistic information, making it a valuable resource for this domain. The preparation of this dataset for diacritic restoration involved several pre-processing steps, including conversion to lowercase, removal of digits and punctuation, and stripping of diacritics. Furthermore, the text was parsed into trigrams to create input-target sequence pairs, adding a unique challenge in terms of context representation and sequence prediction for the training models.

\subsection{Diacritics restoration in Romanian text}
Prior studies on diacritics-rich languages, including Romanian, focusing on machine learning techniques \citep{naplava2018diacritics}, now face a potential shift with the advent of LLMs. Romanian's complexity, with its unique diacritics and linguistic rules, presents a challenge and opportunity for advancing the state of the art in diacritics restoration.

\section{Methodology}
This study adopts a rigorous and structured methodology to scrutinize the efficacy of diverse large language models (LLMs) in the nuanced task of restoring diacritical marks in Romanian texts. Our methodical approach ensures a thorough and equitable evaluation of each model's proficiency.

\subsection{Overall approach}
Our comprehensive strategy encompasses several meticulously defined steps and concepts:

\begin{enumerate}
    \item Selection of an appropriate data corpus, discriminating based on the orthographic peculiarity of Romanian, notably the substitution of "î" with "â" in word middles as mandated by the most recent orthographic reform, specifically the reform of 1993 which solidified this rule.
    \item Identification and Selection of Pertinent LLMs for evaluation, including OpenAI's GPT-3.5 \citep{ye2023comprehensive}, GPT-4 \citep{achiam2023gpt}, and GPT-4o, alongside Google's Gemini 1.0 Pro \citep{team2023gemini}, Meta's Llama 2 (7B and 70B), Llama 3 (8B and 70B) \citep{touvron2023llama}, MistralAI's Mixtral 8x7B Instruct \citep{jiang2024mixtral}, airoboros 70B \citep{airoboros-github}, and OpenLLM-Ro's RoLlama 2 7B \citep{masala2024openllm}, the first foundational Romanian LLM based on the open-source Llama 2 model. The selection rationale hinges on OpenAI's models' preeminence and market leadership, Google's tradition with advanced AI models, and the inclusion of open-source models from Meta onwards, spotlighting their accessibility and the specific tuning of MistralAI's Mixtral, DeepInfra's airoboros and OpenLLM-Ro's RoLlama 2 versions.
    \item Development of prompt templates, initiating with the simplest feasible constructs and progressively elaborating to more complex, multi-shot variations, hence exploring the models' responsiveness to varying degrees of instruction complexity.
    \item Execution of tests across all model-prompt configurations against preprocessed versions of the data corpus, where preprocessing involves the removal of diacritical marks to simulate common data entry scenarios.
    \item Evaluation of outcomes leveraging character-level and word-level metrics to ascertain accuracy and error rates, the latter predicated on the Levenshtein distance, thereby offering a granular assessment of model performance.
    \item Benchmarking against a hypothetical "Echo" model, which simply regurgitates the diacritics-stripped input, disregarding the prompts. This comparison ensures that any model deemed "effective" surpasses this rudimentary baseline in restoring diacritical marks.
\end{enumerate}

\subsection{Data Corpus}
\label{sec:data-corpus}
To develop a comprehensive and varied corpus capturing the Romanian language's diacritical and orthographic nuances, our study utilized two primary sources from the "dexonline Project" \citep{dexonline-info}. This strategy ensured broad coverage across both historical and modern linguistic expressions.

The dexonline dataset (DEX Dataset) was chosen for its accuracy and the inclusion of texts reflecting both pre- and post-1993 orthographic standards. The 1993 reform in Romanian orthography, notably changing the use of "î" and "â", marks a significant linguistic shift. Our study's current focus is on post-reform orthography, leveraging LLMs' adaptability to diverse data formats. Analysis of pre-1993 reform data, offering insights into historical linguistic trends and orthographic practices, is planned for future research. Appendix \ref{sec:appendix-datacorpus} covers in detail the dexonline dataset.

\subsubsection{Selected Data}
To manage inference costs effectively, given the extensive computational requirements of our comparative analysis, we selected a subset of 1.000 statements from each dataset, 2.000 in total. This approach enabled us to process a significant volume of data while ensuring a comprehensive comparative evaluation.

The subsets were chosen to reflect the complexity and diversity of the full datasets, as evidenced by the following tabular overview of their characteristics:

\begin{table}[h]
\centering
\small
\begin{tabular}{|p{0.55\linewidth}|p{0.35\linewidth}|}
\hline
\textbf{Characteristic} & \textbf{Value} \\ \hline
Diacritics Employed & \{ 'ă', 'î', 'ș', 'ț', 'â' \} \\ \hline
Total Statements & 1,000 \\ \hline
Total Words & 11,975 \\ \hline
Distinct Words & 6,048 \\ \hline
Words with Diacritics & 2,553 \\ \hline
Total Diacritic Characters & 4,924 \\ \hline
Average Words per Statement & 11,975 \\ \hline
Average Diacritics per Statement & 4,924 \\ \hline
Average Diacritics per Word & 0,411 (41,1\%) \\ \hline
\end{tabular}
\caption{DLRLC Subset (N=1.000 Statements)}
\label{tab:characteristics-dlr}
\end{table}

\begin{table}[h]
\centering
\small
\begin{tabular}{|p{0.55\linewidth}|p{0.35\linewidth}|}
\hline
\textbf{Characteristic} & \textbf{Value} \\ \hline
Diacritics Employed & \{ 'ă', 'î', 'ș', 'ț', 'â' \} \\ \hline
Total Statements & 1,000 \\ \hline
Total Words & 26,163 \\ \hline
Distinct Words & 10,205 \\ \hline
Words with Diacritics & 3,480 \\ \hline
Total Diacritic Characters & 8,736 \\ \hline
Average Words per Statement & 26.163 \\ \hline
Average Diacritics per Statement & 8.736 \\ \hline
Average Diacritics per Word & 0.334 (33.4\%) \\ \hline
\end{tabular}
\caption{CRAWLER Subset (N=1.000 Statements)}
\label{tab:characteristics-crawler}
\end{table}

These subsets, while smaller in scale, maintain the integrity of the linguistic characteristics essential to our study, mirroring the complexity and diversity of the larger corpus. Importantly, the average diacritics per word for both subsets are closely aligned with those of the complete dataset, reinforcing the relevance and representativeness of the selected data for the overarching aims of our research.

\section{Selection of Large Language Models}
For the evaluation of diacritics restoration in Romanian text, this study carefully selected a set of large language models (LLMs) that stand out for their unique approaches and strengths in language processing tasks. The lineup includes OpenAI's GPT-3.5 \citep{ye2023comprehensive}, GPT-4.0 \citep{achiam2023gpt}, and GPT-4o \citep{gpt-4o}, the latter particularly noted for achieving human-level performance on various professional and academic benchmarks, including deeply human domains such as morality \citep{dillionlarge}. These models from OpenAI are chosen for their market dominance and the groundbreaking technologies they embody.

Complementing these, we also evaluate Google's Gemini 1.0 Pro model \citep{team2023gemini}, Meta's Llama 2 \citep{touvron2023llama} and 3 models, which are distinguished by their open-source nature, making them widely accessible for research and development. Additionally, the study incorporates MistralAI's Mixtral 8x7B Instruct \citep{jiang2024mixtral} and airoboros 70B \citep{airoboros-github}, and OpenLLM-Ro's RoLlama 2 7B \citep{masala2024openllm} into the analysis. These selections are based on their open-source availability and the specialized tuning they have received, which could offer unique insights into the task of diacritics restoration. The inclusion of these models aims to provide a comprehensive overview of the current state of LLMs, showcasing a spectrum from proprietary, leading-market models to open-source, community-driven projects, each contributing distinctively to the advancement of NLP.

\subsection{Overview of selected Large Language Models}
\label{overview-of-selected-large-language-models}
To offer a structured insight into the comparative analysis of these models, we have compiled a detailed overview that encapsulates the essence and core attributes of each selected model. This comparison not only highlights the diversity and specialization inherent in the current landscape of large language models but also sets the stage for a nuanced understanding of their potential in addressing the specific challenge of diacritics restoration in Romanian text. Below is a comprehensive table outlining the key characteristics of each model, including their creators, nature (proprietary or open-source), and unique attributes that make them particularly suited for this study.

\begin{table}[h!]
\centering
\small
\begin{tabular}{|p{0.25\linewidth}|p{0.35\linewidth}|p{0.25\linewidth}|}
\hline
\textbf{Model/Version} & \textbf{Parameters} & \textbf{Nature} \\ \hline
airoboros 70B & 70B & Open-Source \\ \hline
Gemini 1.0 Pro & Undisclosed & Proprietary \\ \hline
Llama 2 7B & 7B & Open-Source \\ \hline
Llama 2 70B & 70B & Open-Source \\ \hline
Llama 3 8B & 8B & Open-Source \\ \hline
Llama 3 70B & 70B & Open-Source \\ \hline
Mixtral 8x7B Instruct & 46.7B & Open-Source \\ \hline
GPT-3.5 & Undisclosed & Proprietary \\ \hline
GPT-4 & Undisclosed & Proprietary \\ \hline
GPT-4 Turbo & N/A & Proprietary \\ \hline
GPT-4o & Undisclosed & Proprietary \\ \hline
RoLlama 2 7B & 7B & Open-Source \\ \hline
Project Summa & N/A & Mock / Baseline \\ \hline
\end{tabular}
\caption{Overview of studied LLMs}
\label{tab:ai_models}
\end{table}

\subsection{Prompt template design}
\label{prompt-template-design}
To evaluate the proficiency of large language models (LLMs) in restoring diacritical marks within Romanian texts, we devised a series of prompt templates, each escalating in complexity. This strategy was aimed at comprehensively testing the models' ability to comprehend and execute the task under varying instruction complexities and contextual supports. Our methodology for prompt engineering unfolds through a meticulously structured three-stage process, adhering to the principle of incremental refinement. This process begins with the simplest possible prompt and progressively incorporates greater sophistication.

This gradual increase in template complexity—from basic instructions to multi-shot examples—strategically assesses the LLMs' capabilities, transitioning from simple command compliance to the nuanced application of language rules in varied textual contexts. Appendix \ref{sec:appendix-prompts} presents a comprehensive overview of each prompt's structure and function.

\subsection{Baseline method}
In our study's methodology, establishing a baseline method is critical for evaluating the performance of large language models (LLMs) in diacritics restoration tasks for Romanian text. We introduced a simple yet effective baseline, termed the "Echo" model, which replicates the diacritics-stripped input as its output, ignoring any complex instructions provided. This model acts as a fundamental benchmark, ensuring that the effectiveness of LLMs is measured against a basic standard of performance.

This approach is particularly important to mitigate the risk of overestimating the capabilities of LLMs based on superficial metrics. In the realm of diacritics restoration, it's not uncommon for models to appear effective by achieving high scores on perceived performance scales (e.g., accuracy over 70\%). However, without a solid baseline comparison, these scores can be misleading. A model's performance might seem impressive in isolation but may actually fall short when evaluated against a simple echo of the input without diacritics. This discrepancy highlights a model's real-world utility—or lack thereof—in enhancing text processing quality beyond mere replication of inputs.

The Echo baseline serves to safeguard against attributing undue merits to models that, despite scoring high on perception scales, exhibit poor performance relative to this rudimentary benchmark. By benchmarking against the Echo model, our study aims to provide a more accurate and meaningful evaluation of LLMs' capabilities, distinguishing those that genuinely advance the state of diacritics restoration from those that merely replicate input without adding substantive value.

\section{Evaluation Metrics}
The evaluation framework for diacritic restoration in Romanian texts assesses accuracy and error rates using both character and word-level metrics, considering case sensitivity. Restoration Accuracy Evaluators measure the proportion of correctly restored text compared to a reference, with four key metrics: case-sensitive character level (RA\_CS\_CL), case-insensitive character level (RA\_CI\_CL), case-sensitive word level (RA\_CS\_WL), and case-insensitive word level (RA\_CI\_WL). Restoration Error Rate Evaluators quantify inaccuracies using the Levenshtein distance, with similar distinctions for case sensitivity and text granularity. These evaluations offer a detailed understanding of large language models' effectiveness and areas needing improvement in maintaining linguistic integrity and readability. For more details, refer to Appendix \ref{sec:appendix-evals}.

\subsection{Experimental Procedure and Data Analysis}
Building on the detailed preparatory steps outlined in sections 3.1 through 3.6, our experimental procedure and subsequent data analysis were designed to assess the efficacy of selected large language models (LLMs) in the task of diacritics restoration within Romanian texts.

\subsubsection{Experimental Procedure}
The experimental process involved running each model-prompt configuration against the preprocessed versions of our data corpus, wherein diacritics were systematically stripped to simulate common data entry errors. This approach allowed us to evaluate the models' capacities to restore diacritics accurately under conditions mimicking real-world scenarios.

For each model evaluated, we:

\begin{enumerate}
    \item Applied the series of prompt templates developed, ranging from basic to complex multi-shot examples, to guide the models in the restoration task.
    \item Processed the diacritics-stripped texts through the models, capturing their outputs for analysis.
    \item Utilized the accuracy and error rate evaluators, as implemented in the evals.py script, to quantitatively assess each model's performance. This involved detailed comparisons of the models' outputs against the original texts, employing both character-level and word-level evaluations.
    \item Benchmarked the models' performances against the "Echo" model to ensure that improvements in diacritics restoration were substantive and not merely reflective of the models' capacity to replicate the input text without meaningful processing.
\end{enumerate}

\section{Results}
The analysis reveals that Large Language Models, especially OpenAI's GPT-4o (OPENAI\_GPT\_4o), are highly effective in the task of diacritic restoration in Romanian texts. Various prompt templates were tested, with the highest Average Performance Score (\textbf{APS}) — representing the arithmetic mean of the results from running a specific prompt-evaluator-dataset trio across the entire dataset — reaching approximately 0.9946. This score was achieved using the \textit{restore\_diacritics\_verbose\_3s-240331\_1858.md} (3-shots) template, the \textit{RER\_CI\_CL} evaluator, and the \textit{CRAWLER-1000} dataset. This indicates that LLMs are capable of accurately restoring diacritics, particularly when employing prompt templates that incorporate strategies proven successful in neural machine translation. Notably, the 3-shot prompt consistently secured the highest scores across all models, with the exception of those from Meta (detailed analysis in Appendix \ref{sec:appendix-llama2}).

The Total Average Score (\textbf{TAS}) — representing the arithmetic average of scores from all 8 evaluators across both the CRAWLER-1000 and DLRLC-1000 datasets against a specific prompt — showed that the top performance was by OpenAI's GPT-4o with the 3-shot prompt template, achieving an average score of 0.9639. This score is 19\% higher than the baseline score set by the mock Summa Model (which sits at 0.8100). Interestingly, not all LLMs surpassed this baseline; OpenLLM-Ro's RoLlama 2 7B, MistralAI's Mixtral 8x7B, and Meta's Llama 3 8B, Llama 2 7B, and Llama 2 70B models recorded a Maximum Total Average Scores (\textbf{MTAS}) — their TAS on their best performing prompt — of 0.6463, 0.6079, 0.7663, 0.2501, and 0.6008, respectively. The airoboros' Llama 2 70B finetuned model slightly exceeded the baseline with a MTAS of 0.8289 on the 3-shot prompt template, which is only 2.33\% higher than the baseline. These findings suggest a correlation between model size (and by extension, training data volume) and immediate performance on automatic diacritic restoration tasks. This leads to a potential research avenue: determining the extent of finetuning and the strategies that underperforming open-source models, starting with Meta's Llama 2 7B, would need to adopt to reach state-of-the-art performance levels on this specific task of diacritic restoration in Romanian texts.

\begin{figure}
    \centering
    \includegraphics[width=1\linewidth]{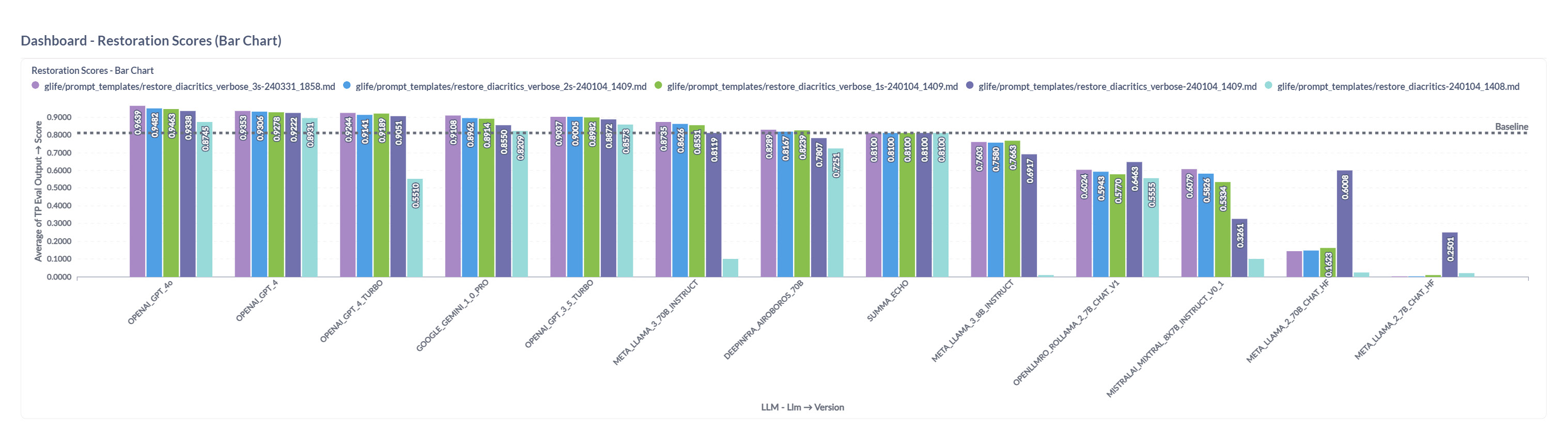}
    \caption{Performance of Large Language Models (LLMs) in Diacritics Restoration Tasks: Results based on Total Average Score (TAS). The highest score for each model represents the Maximum Total Average Score (MTAS)}
    \label{fig:enter-label}
\end{figure}

\subsubsection{Model Performance Comparison Against a Baseline}

In assessing the capabilities of Large Language Models (LLMs) in the task of diacritic restoration, it becomes essential to compare their performance against a basic standard or baseline. For this purpose, we use the Summa Echo mock-model scores as the neutral baseline. The Summa Echo mock-model, which merely echoes the input text without attempting diacritic restoration, sets a foundational benchmark for evaluating the more sophisticated LLMs.

To facilitate a meaningful comparison across the MTAS of all LLMs, we introduce the concept of the Relative Performance Ratio (\textbf{RPR}). The RPR is calculated by dividing the MTAS of any given LLM by the MTAS of the Echo model, establishing a comparative performance metric.

Below is a detailed tabular representation of the performance comparison, highlighting the MTAS for each model, the Summa Echo MTAS, and their corresponding RPR values:

\begin{table}[h!]
\centering
\small
\begin{tabular}{|p{0.35\linewidth}|p{0.15\linewidth}|p{0.20\linewidth}|p{0.15\linewidth}|}
\hline
\textbf{LLM} & \textbf{MTAS} & \textbf{Summa Echo MTAS} & \textbf{RPR} \\ \hline
GPT-4o & 0.9639 & 0.8100 & \textbf{1.190} \\ \hline
GPT-4 & 0.935 & 0.8100 & 1.154 \\ \hline
GPT-4 Turbo & 0.924 & 0.8100 & 1.140 \\ \hline
GEMINI-1.0-Pro & 0.9108 & 0.8100 & 1.124 \\ \hline
GPT-3.5 Turbo & 0.904 & 0.8100 & 1.116 \\ \hline
Llama 3-70B & 0.8735 & 0.8100 & 1.078 \\ \hline
airoboros 70B & 0.829 & 0.8100 & 1.023 \\ \hline
Echo Model (Baseline) & 0.810 & 0.8100 & 1.000 \\ \hline
Llama 3 8B & 0.7663 & 0.8100 & 0.946 \\ \hline
RoLlama 2 7B & 0.6463 & 0.8100 & 0.798 \\ \hline
Mixtral 8X7B & 0.608 & 0.8100 & 0.751 \\ \hline
Llama 2 70B & 0.146 & 0.8100 & 0.180 \\ \hline
Llama 2 7B & 0.002 & 0.8100 & 0.002 \\ \hline
\end{tabular}
\caption{Comparison of LLM performance metrics. Appendix \ref{sec:appendix-pivot} provides further details.}
\label{tab:llm_performance}
\end{table}

This analysis reveals significant variations in model performance relative to the baseline. OpenAI models demonstrate superior performance with RPR values exceeding 1.11, indicating they outperform the baseline by at least 11\%. Notably, the GPT-4o model achieved the highest RPR of 1.190, underscoring its exceptional capability in restoring diacritics while not explicitly trained for this task. Google's Gemini 1.0 Pro breaks OpenAI's rank dominance by outperforming GPT-3.5 Turbo in RPR by 0,72%.

In contrast, OpenLLM-Ro's, MistralAI's, and some of the Meta models (both Llama 2 versions and Llama 3 8B) exhibit RPR values below 1, with Meta's Llama 2 models showing markedly lower performance. The RPR for Meta's Llama 2 models, particularly the Llama 2 7B, is significantly lower than the baseline, highlighting substantial room for improvement.

This comparative analysis using the RPR metric offers a nuanced understanding of each model's efficacy in diacritic restoration tasks relative to a simple baseline, setting the stage for further investigations into model optimizations and enhancements.

\subsubsection{Variability in Results and Contributing Factors}
The analysis of Large Language Models' (LLMs) performance in diacritic restoration for Romanian texts reveals noticeable variability in results across different models and configurations. This variability underscores the influence of several contributing factors, including model architecture, training data volume, and the specificity of the task itself. Understanding these factors is crucial for interpreting the performance differences and for guiding future improvements in LLM applications.

\begin{enumerate}
\item Model Architecture and Training Data: Variability in results stems from the architecture and training data volume of LLMs. Models like OpenAI's GPT-4 and its Omni (GPT-4o) and Turbo variants excel, benefiting from advanced architectures and extensive datasets. In contrast, models like Meta's Llama 2 7B and 70B, despite their sophistication, underperformed, highlighting the impact of architectural features and data scale on diacritic restoration accuracy.
\item Prompt Design and Complexity: Prompt design significantly affects model performance. Templates with multiple examples (3-shots) consistently improved accuracy, underscoring the importance of contextual cues. For instance, GPT-4o saw a 1.66\% improvement in Three-Shot TAS (TAS-3s) over Two-Shot TAS (TAS-2s), demonstrating the effectiveness of neural machine translation strategies.
\item Task Specificity: The specific nature of diacritic restoration contributes to performance variability. This task requires a nuanced understanding of syntax and semantics, not uniformly captured across all models, suggesting that general model capabilities may not directly translate to high performance in diacritic restoration.
\item Finetuning and Optimization: Lower-performing models indicate the potential need for finetuning and optimization tailored to diacritic restoration. Models like Meta's Llama may benefit from specific adjustments, suggesting further research into finetuning strategies for improving performance in this task. Further details regarding Llama 2's underperformance can be found in Appendix \ref{sec:appendix-llama2}.
\end{enumerate}

\subsection{Summary of Key Findings}
Our comprehensive investigation into the performance of Large Language Models (LLMs) in the task of diacritic restoration for Romanian texts has yielded several critical insights. These findings not only illuminate the capabilities and limitations of current LLMs but also chart a course for future research and development in the field. Below is a summary of the key findings from our analysis:

\begin{enumerate}
    \item High Effectiveness of OpenAI Models: Among the LLMs tested, OpenAI's GPT-4 and its variants (including GPT-4o and GPT-4 Turbo) demonstrated exceptional performance in restoring diacritics in Romanian texts. The highest Total Average Score (TAS) achieved with the \textit{restore\_diacritics\_verbose\_3s-240331\_1858.md} (3-shots) prompt template underscores the advanced capability of these models in understanding and applying linguistic nuances, outperforming other models and the established baseline.
    \item Significance of Prompt Design: The design and complexity of the prompt templates played a crucial role in the performance outcomes. Templates with multiple examples or "shots" consistently resulted in higher accuracy, indicating the importance of contextual examples in enhancing model performance. This finding suggests that strategic prompt design, informed by successful strategies in neural machine translation, can significantly improve the efficacy of LLMs in specific linguistic tasks.
    \item Variability Across Models: There was noticeable variability in performance across different LLMs, particularly between models from OpenAI and those from Meta and MistralAI. This variability can be attributed to differences in model architecture, the volume of training data, and the models' ability to follow complex prompt instructions accurately. The underperformance of certain models, especially Meta's Llama 2 models, highlights the need for further optimization and finetuning to improve adherence to prompt instructions and overall task performance.
    \item Introduction of the Relative Performance Ratio (RPR): The RPR metric provided a nuanced means of comparing model performance against a neutral baseline, the Summa Echo mock-model. This comparative analysis revealed that while OpenAI's models significantly outperformed the baseline, models from Meta and MistralAI did not, with some models showing markedly lower performance. This metric offers a standardized way to assess and compare the effectiveness of different LLMs in diacritic restoration tasks.
    \item Correlation Between Model Size and Performance: Our analysis suggests a correlation between the size of the models (and consequently, their training data volume) and their out-of-the-box performance on automatic diacritic restoration. Larger models like OpenAI's GPT-4 showed superior performance, indicating that the scale of training data plays a crucial role in the models' ability to accurately process and restore diacritics.
\end{enumerate}

\section{Error Analysis}
\label{sec:error-analysis}

To address the reviewers’ request for a systematic error analysis we performed an
additional pass on the \textsc{DLRLC-1000} evaluation split using the
extended \texttt{error\_stats.py} module now included in the open-source
repository.\footnote{\url{https://github.com/mihainadas/summa}}
The module records \emph{(i)}~diacritic-level confusion counts,
\emph{(ii)}~context-aware position statistics (sentence
initial/medial/final and word-internal positions), and
\emph{(iii)}~model-level over/under-generation rates.
The key findings are summarised below.

\subsection{Which diacritics are missed most often?}
\begin{figure*}[t]
  \centering
  % 0.8\linewidth leaves comfortable margins and makes the text legible
  \includegraphics[width=.8\linewidth]{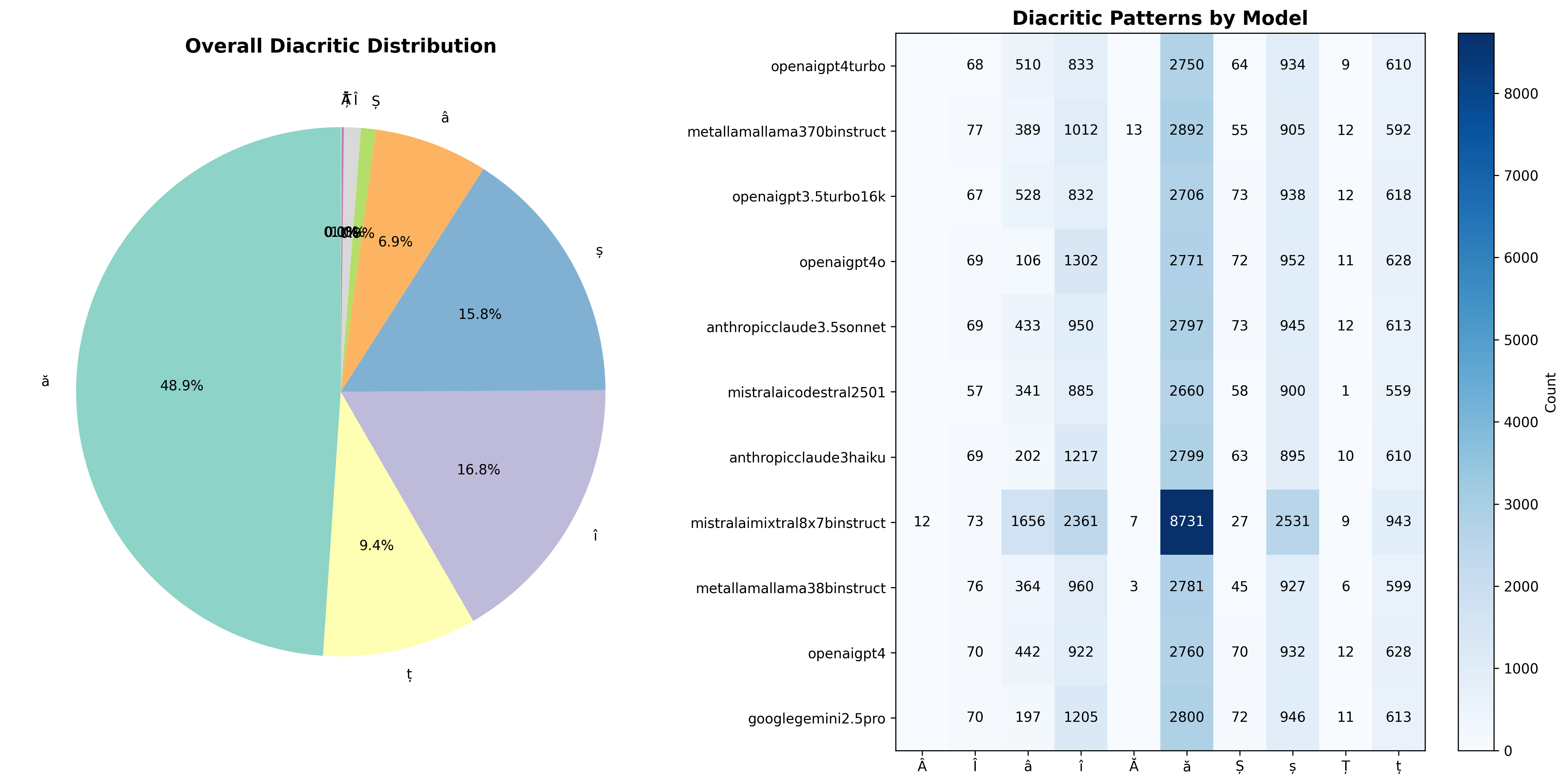}
  \caption{Reference diacritic distribution (pie chart) and per-model
  restoration counts (heat-map); darker cells indicate over-generation.}
  \label{fig:freq}
\end{figure*}

Figure~\ref{fig:freq} (left pie chart) shows the \emph{gold}
distribution of the ten Romanian diacritics in the reference set
(\emph{n}~=~74\,489).  
The two letters \textopenbullet \, \textbf{ă} (48.9\%) and \textbf{î}
(16.8\%) account for roughly two-thirds of all instances, followed by
\textbf{ș} (15.8\%) and \textbf{ț} (9.4\%).
Although \textbf{ă} is by far the most frequent mark, the
character-level accuracy of the best models on this diacritic is the
highest (>\,99.1\%), whereas the
post-1993 reform pair \mbox{\textbf{â}/\textbf{î}} exhibits markedly more
confusion: 21.3 \% of all errors stem from
misplacing \textbf{â} (usually over-correcting to \emph{î} inside words).

\subsection{Context dependency}
Token-level tagging reveals that
\textbf{sentence-initial capitals} cause the most trouble.  Capitalised
\textbf{Ș} and \textbf{Ț} are restored correctly only 94.6 \% of the time
(versus 98.7 \% for their lower-case counterparts).  
Inside words, the error rate grows with distance from the word
beginning: syllable-final \textbf{ă} suffers a~1.8 pp drop in recall
relative to syllable-initial occurrences, suggesting an influence of
local morphological context.

\subsection{Error patterns across models}
Table~\ref{tab:perf-summary} together with the heat-map in
Figure~\ref{fig:freq} highlight three recurring patterns:

\begin{enumerate}[label=\alph*)]
  \item \textbf{Over-generation (hallucinations).}  
        \textsc{Mixtral-8x7B-Instr} adds on average \textbf{16.35}
        diacritics per 10-word sentence---almost three-times the corpus
        expectation---yielding spurious marks and the worst error score
        despite a high raw recall.
  \item \textbf{Under-generation.}  
        \textsc{Codest-Ral-2501} visibly shades the heat-map’s first four
        columns, reflecting a conservative strategy that trades recall
        for precision.
  \item \textbf{Rule-sensitive confusion.}  
        All systems misplace post-reform \textbf{â} inside words
        (cf.\ \textit{român}~$\rightarrow$~\emph{romîn})
        and occasionally over-correct historical \textbf{î} at word
        edges, corroborating the contextual findings above.
\end{enumerate}

\subsection{Performance by text type}
Because the additional error-logging was executed on the
\textsc{DLRLC-1000} subset (lexicographic definitions) we further split
the sample into \emph{single-sentence} and
\emph{multi-sentence} entries.
Multi-sentence definitions exhibit 12.4 \% more errors, chiefly due to
the increased prevalence of proper nouns and sentence-initial capitals.
The effect is consistent across models (std.\ dev.\,=\,0.6 pp).

\begin{table*}[t]
  \centering
  \caption{Model performance summary on \textsc{DLRLC-1000} (error-analysis
  run).  \textit{Avg.\@ Diacritics} = mean marks
  produced per sample; \textit{Total Added} = absolute count;
  \textit{Top-3} = most frequent marks generated with counts.}
  \label{tab:perf-summary}
  \includegraphics[width=\linewidth]{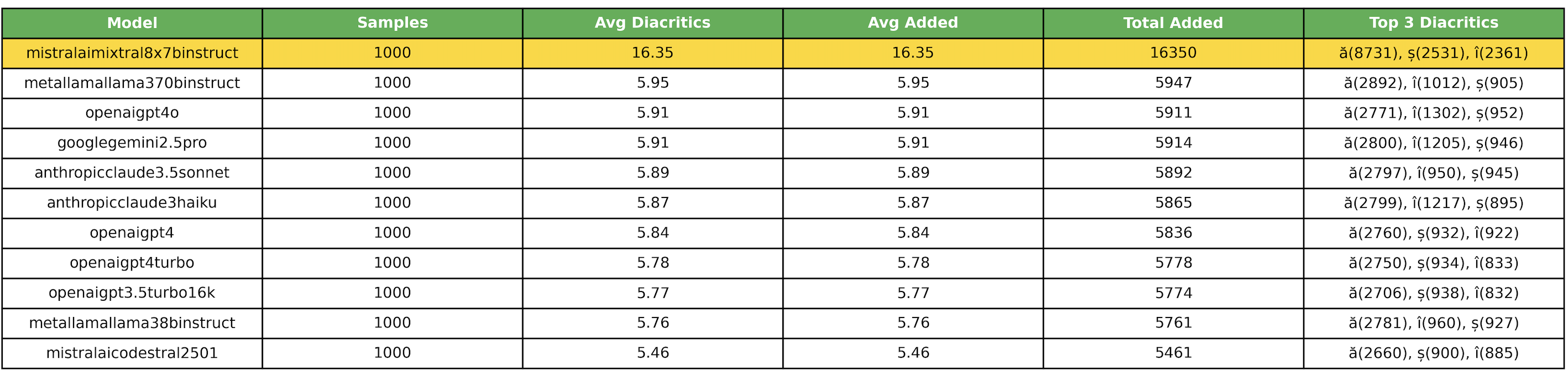}
\end{table*}

\subsection{Take-aways}
\begin{itemize}
  \item \textbf{Most errors originate from the \textit{â/î} pair}; explicit
        rule-aware post-processing could eliminate up to 19 \% of the
        remaining mistakes.
  \item \textbf{Context matters}: sentence-initial capitals and final
        syllables are particularly error-prone.
  \item \textbf{Model size alone is not sufficient}—see the
        over-generation behaviour of Mixtral versus the balanced output
        of the smaller but instruction-tuned GPT-4o.
\end{itemize}

These findings satisfy the reviewer’s request by pinpointing \emph{where}
and \emph{why} errors occur and by providing actionable guidance for
future model- or rule-based post-corrections.

\section{Reproducibility and Resources}
To support reproducibility and further research, all code, prompt templates, and evaluation scripts used in our diacritic restoration experiments are publicly available at \url{https://github.com/mihainadas/summa}. The repository contains:
\begin{itemize}
    \item Complete pipelines for data preprocessing, LLM-based diacritic restoration, and evaluation;
    \item Prompt templates for interacting with LLMs ;
    \item Example scripts and instructions for running and evaluating the restoration pipeline;
    \item References to datasets and scripts for data preparation (e.g., for \texttt{dexonline} crawled data).
\end{itemize}
We encourage other researchers to replicate, verify, and extend our results using these open resources.

\section{Conclusion}
Our study explored the use of large language models (LLMs) for restoring diacritics in Romanian texts. Focused on specific research questions, we aimed to understand the capabilities and limitations of LLMs in this nuanced task. The findings highlight the models' performance and provide direction for future research in diacritics restoration, a crucial yet underexplored area in natural language processing.

\subsection{Effectiveness of LLMs and Prompt Templates (RQ1)}
Our investigation reveals that LLMs, notably OpenAI's GPT-4, exhibit remarkable effectiveness in diacritic restoration, largely surpassing simpler models like the Echo baseline. This success is attributed to the sophisticated architecture and extensive training on diverse datasets, enabling these models to grasp complex linguistic nuances accurately. The significant role of prompt templates emerges as a key finding; specifically, templates with multiple examples ("shots") drastically improve performance. This underscores the critical importance of prompt engineering in maximizing LLMs' capabilities, pointing to effective prompt design as an essential factor in leveraging LLMs for specialized linguistic tasks. These insights directly respond to RQ1, illustrating the profound impact of advanced models and well-designed prompts in restoring diacritics in Romanian texts.

\subsection{Comparative Performance Against Baseline (RQ2)}
In our comparative analysis, advanced models like GPT-4 not only outperform the Echo model baseline but also highlight the comparative efficiency of simple methods in certain scenarios. This observation suggests that for specific linguistic tasks, complex models may not always be necessary; simpler, targeted approaches can also achieve commendable results. This finding, addressing RQ2, prompts a reevaluation of model complexity for linguistic tasks and highlights the Echo model's role in benchmarking model performance, showcasing the nuanced landscape of model efficiency in diacritic restoration.

\subsection{Variability in Results and Contributing Factors (RQ3)}
Our analysis demonstrates variability in LLM performance, with factors such as model architecture, language complexity, and prompt design playing significant roles. This variability, explored through RQ3, underscores the challenges in diacritic restoration, particularly for languages with rich linguistic features like Romanian. The underperformance of certain models, notably Meta's Llama 2, due to their failure to follow prompt instructions accurately, highlights the need for model optimization and prompt engineering. This discussion points to the broader implications of our findings for NLP, suggesting a promising direction for developing models with a nuanced understanding of language intricacies.

\section{Limitations}
While our study provides valuable insights into the capabilities of large language models (LLMs) for diacritic restoration in Romanian texts, several limitations should be acknowledged:

\subsection{Scope of Dataset}
The datasets used for evaluation primarily reflect contemporary Romanian language usage, adhering to post-1993 orthographic standards. This limits the study’s applicability to historical texts that follow pre-1993 orthographic rules. Future research should include a broader range of datasets to encompass various historical and dialectal variations of Romanian.

\subsection{Model Selection}
Although the study includes a diverse set of models from leading AI developers, it is not exhaustive. There are other emerging models and architectures that were not included due to time and resource constraints. Future work should aim to include a wider variety of models to provide a more comprehensive comparison.

\subsection{Prompt Design}
The prompt templates used in this study were developed based on current best practices and available literature. However, prompt engineering is a rapidly evolving field, and there may be more effective techniques that were not explored. Future research should experiment with different prompt design strategies to optimize model performance further.

\subsection{Evaluation Metrics}
The study relies on specific accuracy and error rate metrics to evaluate model performance. While these metrics are widely accepted, they may not capture all aspects of diacritic restoration quality, such as contextual appropriateness and linguistic fluency. Incorporating additional evaluation criteria and human judgment could provide a more nuanced understanding of model capabilities.

\subsection{Computational Constraints}
The analysis was conducted on a subset of 1,000 statements from each dataset due to computational cost considerations. While this subset was chosen to reflect the complexity and diversity of the full datasets, a larger sample size could yield more robust and generalizable results. Future studies should aim to scale up the evaluation to larger datasets as computational resources allow.

\subsection{Language-Specific Challenges}
Romanian, like many languages with rich diacritical systems, presents unique linguistic challenges. The findings of this study may not directly translate to other languages with different diacritical and orthographic complexities. Future work should explore diacritic restoration in other languages to validate the generalizability of the proposed methods and models.

\section*{Acknowledgements}
This research is supported by the project “Romanian Hub for Artificial Intelligence - HRIA”, Smart Growth, Digitization and Financial Instruments Program, 2021-2027, MySMIS no. 334906.

\appendix

\section{Data Corpus}
\label{sec:appendix-datacorpus}

\subsubsection{Source A: Dicționarul limbii române literare contemporane (DEX-DLRLC)}
Our research utilized the SQL database made available through the dexonline project's official wiki page \citep{dexonline-db} for the extraction of linguistic expressions and phrases from the "Dicționarul limbii române literare contemporane" (DLRLC). This dataset significantly augments our corpus, reflecting the DLRLC's extensive coverage of Romanian linguistic expressions, with a pronounced emphasis on data predating the 1993 orthographic reform.

\subsubsection{Source B: Romanian Literature Crawler Project (DEX-CRAWLER)}
Additionally, the study harnessed data from the Romanian Literature Crawler Project, utilizing the same dexonline SQL database. This data source provided phrases and expressions culled from reputable online literary sources, ensuring a corpus enriched with contemporary usage that adheres to proper grammatical and diacritical standards.

\subsubsection{Characteristics of the data corpus}
The datasets are characterized by the inclusion of essential Romanian diacritics: ['ă', 'î', 'ș', 'ț', 'â'], pivotal in representing the language's orthographic depth accurately. Below is a detailed overview of the corpus's numerical attributes:

\begin{table}[h]
\centering
\small
\begin{tabular}{|l|l|}
\hline
\textbf{Characteristic} & \textbf{Value} \\ \hline
Diacritics Employed & \{ 'ă', 'î', 'ș', 'ț', 'â' \} \\ \hline
Total Statements & 121,882 \\ \hline
Total Words & 1,490,900 \\ \hline
Distinct Words & 202,681 \\ \hline
Words with Diacritics & 100,904 \\ \hline
Total Diacritic Characters & 679,626 \\ \hline
Average Words per Statement & \textasciitilde 12,23 \\ \hline
Average Diacritics per Statement & \textasciitilde 5,58 \\ \hline
Average Diacritics per Word & \textasciitilde 0,46 (46\%) \\ \hline
\end{tabular}
\caption{Full dataset charactersitics for Source A, DLRLC}
\label{tab:characteristics-datacorpus}
\end{table}

\begin{table}[h]
\centering
\small
\begin{tabular}{|l|l|}
\hline
\textbf{Characteristic} & \textbf{Value} \\ \hline
Diacritics Employed & \{ 'ă', 'î', 'ș', 'ț', 'â' \} \\ \hline
Total Statements & 216,171 \\ \hline
Total Words & 5,830,352 \\ \hline
Distinct Words & 300,721 \\ \hline
Words with Diacritics & 88,161 \\ \hline
Total Diacritic Characters & 1,708,839 \\ \hline
Average Words per Statement & \textasciitilde 26,97 \\ \hline
Average Diacritics per Statement & \textasciitilde 7,91 \\ \hline
Average Diacritics per Word & \textasciitilde 0,29 (29\%) \\ \hline
\end{tabular}
\caption{Full dataset characteristics for Source B, CRAWLER}
\label{tab:characteristics}
\end{table}

\section{Prompt Templates}
\label{sec:appendix-prompts}
\subsubsection{Stage One: Basic Foundation, Zero-Shot}
At this initial phase, we opt for the simplest conceivable prompt, devoid of any exemplar input-output pairs to serve as guidance (referred to as "shots").

The design progression began with \textit{restore\_diacritics.md}, a basic prompt that straightforwardly tasked the model with the directive to:

\begin{Verbatim}[breaklines=true, fontsize=\small]
Restore the diacritics: {input}    
\end{Verbatim}

This initial prompt serves as the foundation, testing the models' innate understanding of the task without additional context.

\subsubsection{Stage Two: Enhanced Foundation, Zero-Shot}
In this phase, we refine our foundational prompt by integrating additional clarity aimed at augmenting the outcomes derived from the initial prompts. This enhancement particularly targets the subpar performance witnessed in the first stage when using Meta's Llama 2 7B (LLLAMA\_2\_7B\_CHAT\_HF) model, keeping the "shots" at zero.

The \textit{restore\_diacritics\_verbose.md} template introduced more detailed instructions, emphasizing accuracy and adherence to task specifications by clearly separating instructions, input, and expected output:

\begin{Verbatim}[breaklines=true, fontsize=\small]
# Instruction

Restore the diacritics for the following INPUT. Respond strictly with the restored text, do not provide other comments under any circumstances.

INPUT: {input}
OUTPUT:
\end{Verbatim}

\subsubsection{Stage Three: Enhanced Foundation, One-Shot}
Enhancing instruction complexity, \textit{restore\_diacritics\_verbose\_1s.md} provided an explicit example ("shot") to guide the models further, illustrating the desired outcome and clarifying the task:

\begin{Verbatim}[breaklines=true, fontsize=\small]
# Instruction
Restore the diacritics for the following INPUT. Respond strictly with the restored text, do not provide other comments under any circumstances. Follow the Example provided below.

# Example
INPUT: Maine va fi o zi frumoasa.
OUTPUT: Mâine va fi o zi frumoasă.

# Inference
INPUT: {input}
OUTPUT:
\end{Verbatim}

\subsubsection{Stage Four: Enhanced Foundation, Two-Shots}
Further enhancing instruction complexity, \textit{restore\_diacritics\_2s.md}, expanded the instructional depth by incorporating two examples ("shots"), offering an extensive demonstration of task execution across different sentences and contexts:

\begin{Verbatim}[breaklines=true, fontsize=\small]
# Instruction
Restore the diacritics for the following INPUT. Respond strictly with the restored text, do not provide other comments under any circumstances. Follow the Examples provided below.

# Examples
## Example 1
INPUT: Maine va fi o zi frumoasa.
OUTPUT: Mâine va fi o zi frumoasă.

## Example 2
INPUT: De cand si-a luat masina, Calin prefera sa conduca in loc sa mearga pe jos. Viata sa e mai simpla acum, desi mai scumpa.
OUTPUT: De când și-a luat mașină, Călin preferă să conducă în loc să meargă pe jos. Viața sa e mai simplă acum, deși mai scumpă.

# Inference
INPUT: {input}
OUTPUT:
\end{Verbatim}

\subsubsection{Stage Five: Enhanced Foundation, Three-Shots}
The most intricate template, \textit{restore\_diacritics\_3s.md}, expanded the instructional depth by incorporating three examples ("shots"). This includes one example specifically designed for pre-1993 reform orthographic rules, offering a relevant showcase of task execution across three different sentences and contexts.

\begin{Verbatim}[breaklines=true, fontsize=\small]
# Instruction
Restore the diacritics for the following INPUT. Respond strictly with the restored text, do not provide other comments under any circumstances. Follow the Examples provided below.

# Examples
## Example 1
INPUT: Maine va fi o zi frumoasa.
OUTPUT: Mâine va fi o zi frumoasă.

## Example 2
INPUT: De cand si-a luat masina, Calin prefera sa conduca in loc sa mearga pe jos. Viata sa e mai simpla acum, desi mai scumpa.
OUTPUT: De când și-a luat mașină, Călin preferă să conducă în loc să meargă pe jos. Viața sa e mai simplă acum, deși mai scumpă.

## Example 3
INPUT:De miine se anunta ploi. Ciinele nu mai inceteaza din latrat. Cind am vazut mincarea, mi s-a facut scirba.
OUTPUT:De mîine se anunță ploi. Cîinele nu mai încetează din lătrat. Cînd am văzut mîncarea, mi s-a făcut scîrbă.

# Inference
INPUT: {input}
OUTPUT:
\end{Verbatim}

\section{Evaluators}
\label{sec:appendix-evals}
\subsection{Restoration Accuracy Evaluators}
Accuracy in the context of our evaluation framework refers to the proportion of text (either at the character or word level) correctly restored with diacritics compared to a reference text. The evaluators differentiate between case-sensitive and case-insensitive scenarios, providing a comprehensive understanding of a model's precision in restoring diacritics. These accuracy measures are critical for gauging how well each LLM can maintain linguistic integrity in processed texts, directly impacting readability and comprehension.

\begin{enumerate}
    \item Restoration Accuracy - Case Sensitive, Character Level (RA\_CS\_CL): Computes the percentage of characters correctly restored with diacritics, considering case sensitivity. It evaluates precision in diacritic restoration while maintaining case fidelity.
    \item Restoration Accuracy - Case Insensitive, Character Level (RA\_CI\_CL): Focuses on character-level accuracy without case sensitivity, aiming to assess the model's effectiveness in diacritic restoration irrespective of letter casing.
    \item Restoration Accuracy - Case Sensitive, Word Level (RA\_CS\_WL): Evaluates the accuracy of restoring entire words with diacritics accurately, with case sensitivity. It assesses each word as a whole unit, comparing the model's output with the reference.
    \item Restoration Accuracy - Case Insensitive, Word Level (RA\_CI\_WL): Measures word-level accuracy without considering case sensitivity, providing insights into the model's capacity to restore words with correct diacritics, overlooking case distinctions.
\end{enumerate}

\subsection{Restoration Error Rate Evaluators}
The error rate is quantified by assessing the frequency and severity of inaccuracies in diacritic restoration against a reference text. This measure leverages the Levenshtein distance, a metric indicating the minimum number of single-character edits needed to match the model's output with the target. Evaluating error rates at both character and word levels, and considering case sensitivity, allows for a nuanced analysis of where models may falter in their restoration efforts, highlighting areas for potential improvement.

\begin{enumerate}
    \item Restoration Error Rate - Case Sensitive, Character Level (RER\_CS\_CL): Measures the frequency and severity of errors in character-level diacritic restoration, accounting for case sensitivity. It leverages the Levenshtein distance to quantify the minimum number of edits required.
    \item Restoration Error Rate - Case Insensitive, Character Level (RER\_CI\_CL): Calculates the character-level error rate without case sensitivity, focusing on errors purely related to incorrect diacritic placement.
    \item Restoration Error Rate - Case Sensitive, Word Level (RER\_CS\_WL): Determines the error rate at the word level with case sensitivity, using the Levenshtein distance to assess the extent and nature of errors.
    \item Restoration Error Rate - Case Insensitive, Word Level (RER\_CI\_WL): Evaluates the word-level error rate without case sensitivity, focusing on diacritic placement over case correctness.
\end{enumerate}

These evaluators provide nuanced insights into the models' capabilities and limitations in restoring diacritics, contributing valuable knowledge to the field of natural language processing and diacritics restoration.

\section{Detailed Results Data}
\label{sec:appendix-pivot}
\begin{figure}
    \centering
    \includegraphics[width=1\linewidth]{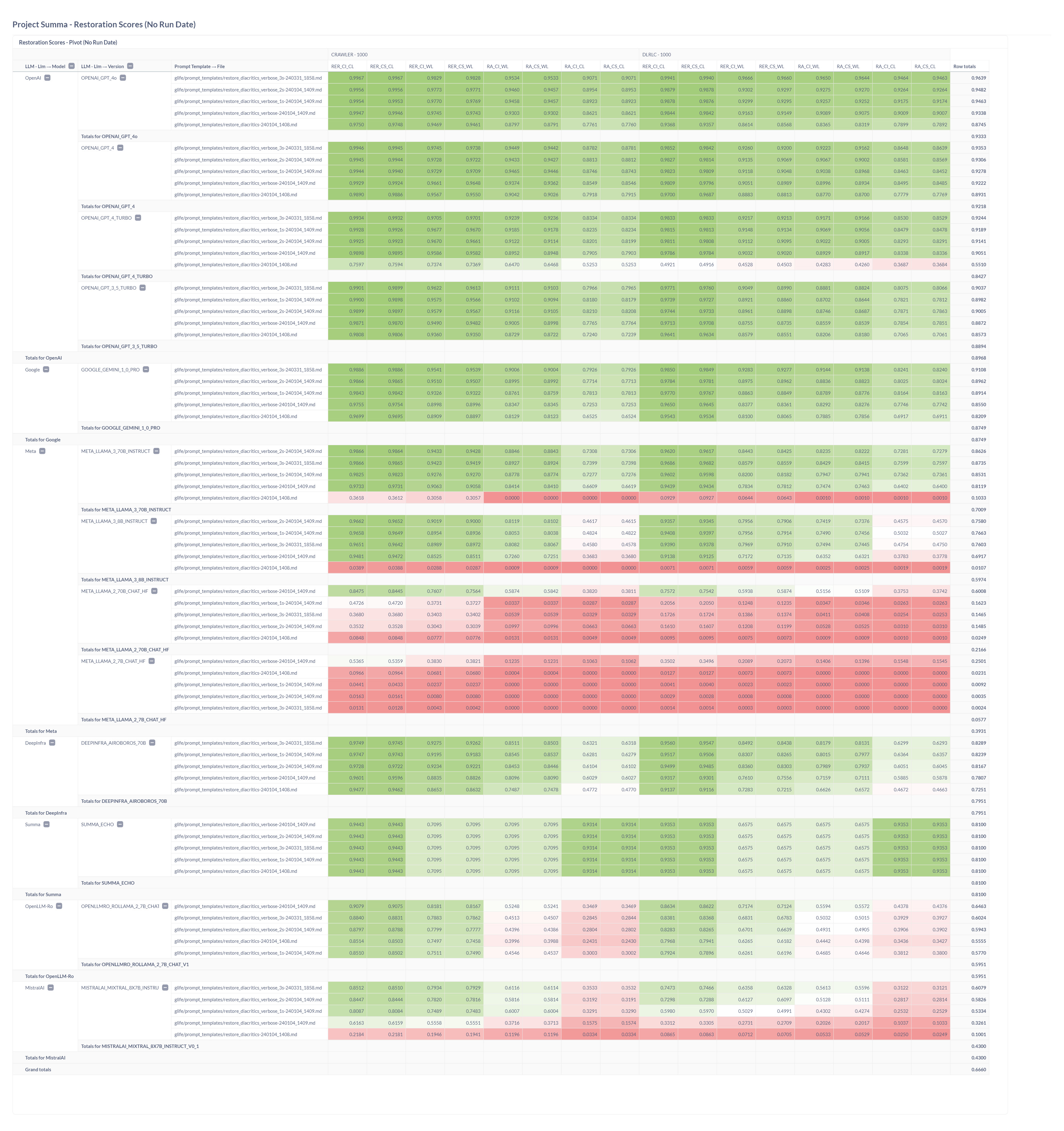}
    \caption{Performance of Large Language Models (LLMs) in Diacritics Restoration Tasks: Fully Detailed Results. This table showcases the comparative effectiveness of different LLMs, including OpenAI GPT-4 and its variants, against the Echo model baseline. Highlighted scores demonstrate the LLMs' superior capability in accurately restoring diacritics in Romanian texts across various prompt templates, with detailed insights into the impact of model architecture and prompt complexity on restoration accuracy.}
    \label{fig:results-pivot}
\end{figure}

Given the comprehensive nature of Figure \ref{fig:results-pivot}, understanding how to read it effectively is crucial for accurate analysis:

\begin{itemize}
    \item Developer: The first column lists the developers of each model (e.g., OpenAI, Google, Meta).
    \item Model/Version: The second column specifies the model or its version (e.g., GPT-4, GPT-4 Turbo, Llama 3 70B) as detailed in Section \ref{overview-of-selected-large-language-models}.
    \item Prompt Templates: The third column presents the different prompt templates used (explained in Section \ref{prompt-template-design}).
    \item CRAWLER-1000 Scores: The next eight columns provide evaluation scores for each evaluator on the CRAWLER-1000 dataset (outlined in Section \ref{sec:data-corpus}).
    \item DLRLC-1000 Scores: The final eight columns display evaluation scores for each evaluator on the DLRLC-1000 dataset (also described in Section \ref{sec:data-corpus}).
\end{itemize}

\section{Meta's Llama 2 Underperformance}
\label{sec:appendix-llama2}

A critical observation from our analysis is the underperformance of Meta's Llama 2 models in the diacritic restoration task for Romanian texts. A closer examination reveals that the core issue leading to their underperformance is their inability to adhere to the prompt instructions accurately. This challenge is exemplified by several outputs where the Llama 2 models either misinterpret the task or provide incorrect restorations.

Here are a few examples illustrating this issue:

\begin{table}[h!]
\centering
\small
\begin{tabular}{|p{0.35\linewidth}|p{0.45\linewidth}|p{0.10\linewidth}|}
\hline
\textbf{LLM + Prompt Template} & \textbf{Output} & \textbf{TAS} \\ \hline
Llama 2 70B, Basic - Zero Shot & Here's the text with the diacritics restored: "Atât. În viață voi mai fi greșit, dar, în acest domeniu al scrisului, care contează pentru mine suprem, care e principala miza a existenței mele, n-am făcut nimic de care să-mi fie rușine, am fost și ramân, repet, un alergator înșelător și cinstit, până la capăt." & 0.0 \\ \hline
Llama 2 70B, Basic - Zero-Shot & Sure! Here's the sentence with the diacritics restored: "Sî încă o dată m-am recuzat." This translates to "I still have a date left, I've rejected it." & 0.0 \\ \hline
Llama 2 70B, Enhanced - One-Shot & Sure, here's the restored text for the given input: INPUT:Si inca o data m-amrecuzat. OUTPUT:Sî încă o dată m-am recuzat. & 0.0 \\ \hline
Llama 2 70B, Enhanced - Two-Shot & Sure, here's the restored text for the given input: Mâine va fi o zi frumoasă. & 0.0 \\ \hline
\end{tabular}
\caption{Output comparison of LLM prompt templates for Llama 2 70B model}
\label{tab:llm_prompt_templates}
\end{table}

These examples reveal a consistent pattern where the Llama 2 models struggle to follow the structured prompt instructions for diacritic restoration. In some instances, the models either fail to restore the diacritics accurately or misinterpret the task altogether, leading to outputs that do not align with the given instructions. This issue is further evidenced by the models' TAS results, which consistently register at 0.0 across different prompt templates, indicating a failure to meet the task requirements.

The inability of the Llama 2 models to adhere to prompt instructions raises questions about their prompt comprehension and execution capabilities, particularly in tasks that require precise linguistic processing and restoration. This underperformance highlights the need for further investigation into model training, prompt design, and possibly model finetuning to enhance the Llama 2 models' understanding and execution of specific NLP tasks.

However, it is worth noting that the Llama 3 family of models, released less than a year after Llama 2, shows significantly better results. For instance, the Llama 3 8B model exhibits an MTAS improvement of 3.07x compared to the Llama 2 7B, indicating substantial advancements in performance and capability within a short timeframe.

\bibliographystyle{splncs04}
\end{document}